\documentclass[sn-basic]{sn-jnl}

\usepackage{lscape}
\usepackage{calc}
\usepackage{paralist}
\newlength{\mylen}
\newcommand{\details}[2]{%
\makebox[\mylen][l]{#1} \,: #2 \\
}
\newcommand{\piped}[1]{%
\text{\textbar}#1\text{\textbar}
}

\jyear{2021}%

\theoremstyle{thmstyleone}%
\theoremstyle{thmstyletwo}%

\theoremstyle{thmstylethree}%

\begin{document}

\title[Automated Layer
Caching]{Improving the Performance of DNN-based Software Services using Automated Layer Caching}


\author*[1]{\fnm{Mohammadamin} \sur{Abedi}}

\author[1]{\fnm{Yani} \sur{Ioannou}}

\author[2]{\fnm{Pooyan} \sur{Jamshidi}}

\author[1]{\fnm{Hadi} \sur{Hemmati}}

\affil[1]{\orgdiv{Electrical and Software Engineering}, \orgname{University of Calgary}, \orgaddress{\country{Canada}}}

\affil[2]{\orgdiv{Computing Science and Engineering}, \orgname{University of South Carolina}, \orgaddress{ \country{United States}}}



\abstract{Deep Neural Networks (DNNs) have become an essential component in many application domains including web-based services. A variety of these services require high throughput and (close to) real-time features, for instance, to respond or react to users' requests or to process a stream of incoming data on time. However, the trend in DNN design is toward larger models with many layers and parameters to achieve more accurate results. Although these models are often pre-trained, the computational complexity in such large models can still be relatively significant, hindering low inference latency. Implementing a caching mechanism is a typical systems engineering solution for speeding up a service response time. However, traditional caching is often not suitable for DNN-based services.
In this paper, we propose an end-to-end automated solution to improve the performance of DNN-based services in terms of their computational complexity and inference latency. Our caching method adopts the ideas of self-distillation of DNN models and early exits. The proposed solution is an automated online layer caching mechanism that allows early exiting of a large model during inference time if the cache model in one of the early exits is confident enough for final prediction. One of the main contributions of this paper is that we have implemented the idea as an online caching, meaning that the cache models do not need access to training data and perform solely based on the incoming data at run-time, making it suitable for applications using pre-trained models.
Our experiments results on two downstream tasks (face and object classification) show that, on average, caching can reduce the computational complexity of those services up to 58\% (in terms of FLOPs count) and improve their inference latency up to 46\%  with low to zero reduction in accuracy.}




\maketitle

\section{Introduction}\label{sec:introduction}
Deep Neural Networks (DNNs) are incorporated in real-world applications used by a broad spectrum of industry sectors including healthcare \citep{Shorten2021DeepLA,Fink2020PotentialCA}, finance \citep{Huang2020DeepLI, Culkin2017MachineLI}, self-driving vehicles \citep{Swinney2021UnmannedAV}, and cybersecurity \citep{Ferrag2020DeepLF}. These applications utilize DNNs in various fields such as computer vision \citep{Hassaballah2020DeepLI,Swinney2021UnmannedAV}, audio signal processing \citep{Arakawa2019ImplementationOD,Tashev2017DNNbasedCV},and natural language processing \citep{Otter2021ASO}.
Many services in large companies such as Google and Amazon have DNN-based back-end software (e.g., Google Lens and Amazon Rekognition) with tremendous volume of queries per second. For instance, Google processes over 99,000 searches every second \citep{mohsin_2022} and spends a substantial amount of computation power and time at their models' run-time \citep{Xiang2019PipelinedDC}. These services are often time-sensitive, resource-intensive, and require high availability and reliability.

Now the question is how fast the current state-of-the-art (STOA) DNN models are at inference time and to what extent they can provide low latency responses to queries. The SOTA model depends on the application domain and the problem at hand. However, the trend in DNN design is indeed toward pre-trained large-scale models due to their reduced training cost (only fine-tuning) while providing dominating results (since they are huge models trained on an extensive dataset).

One of the downsides of large-scale models (pre-trained or not) is their high inference latency. Although the inference latency is usually negligible per instance, as discussed, a relatively slow inference can jeopardize a service's performance in terms of throughput when the QPS is high.

In general, in a DNN-based software development and deployment pipeline, the inference stage is part of the so called ``model serving'' process, which enables the model to serve inference requests or jobs \citep{Xiang2019PipelinedDC} by directly loading the model in the process or by employing serving frameworks such as TensorFlow Serving \citep{Olston2017TensorFlowServingFH} or Clipper \citep{Crankshaw2017ClipperAL}.

The inference phase is an expensive stage in a deep neural model's life cycle in terms of time and computation costs \citep{Desislavov2021ComputeAE}. Therefore, efforts towards decreasing the inference cost in production have increased rapidly throughout the past few years. 

From the software engineering perspective, caching is a standard practice to improve software systems performance, which helps avoid redundant computations. 
Caching is the process of storing recently observed information to be reused when needed in the future, instead of re-computation \citep{Wessels-2001,caching-def}.
Caching is usually orthogonal to the underlying procedure, meaning that it is applied by observing the inputs and outputs of the target procedure and does not engage with the internal computations of the cached function. 

Caching effectiveness is best observed when the cached procedure often receives duplicated inputs while in a similar internal state---for instance, accessing a particular memory block, loading a web page, or fetching the books listed in a specific category in a library database. It is also possible to adopt a standard caching approach with DNNs (e.g., some work cache a DNN's output solely based on its input values \citep{Crankshaw2017ClipperAL}). However, it would most likely provide a meager improvement due to the high dimension and size of the data (such as images, audios, texts) and low duplication among the requests. 

However, due to the feature extracting nature of the deep neural networks, we can expect the inputs with similar outputs (e.g.,\ images of the same person or the same object) to have a pattern in the intermediate layers' activation values. Therefore, we exploit the opportunity to cache a DNN's output based on the intermediate layers' activation values. 
This way, \textbf{we can cache the results not by looking at the raw inputs but by looking at their extracted features in the intermediate layers within the model's forward-pass}.  

The intermediate layers often have even higher dimensions than the input data. Therefore, we use shallow classifiers \citep{Kaya2019ShallowDeepNU} to replace the classic cache storing and look-up procedures. A shallow classifier is a supplementary model attached to an intermediate layer in the base model that uses the intermediate layer's activation values to infer a prediction. In the caching method, training a shallow classifier on a set of samples mimics the procedure of storing those samples in a cache storage, and inferring for a new sample using the shallow classifier mimics the look-up procedure.

Caching is more problematic in regression models where the outputs are continuous values. Specifically, it is less likely that two different samples have the same outcome in a regression model compared to a classification one. Therefore, the experiments in this research focus on classification models.
Thus, here we propose caching the predictions made by off-the-shelf classification models using shallow classifiers trained using the samples and information collected at inference time.

We first evaluate the rationality of our method in our first research question by measuring how it affects the final accuracy of the given base models and assessing the effectiveness of the parameters we introduce (tolerance and confidence thresholds) as a knob to control the caching certainty.

We further evaluate the method in terms of computational complexity and inference latency improvements in the second and third research questions. We measure this improvements by comparing the FLOPs count, memory consumption, and inference latency for the original model vs. the cache-enabled version that we build throughout this experiment. We observed up to 58\% reduction in FLOPs, up to 46\% acceleration in inference latency while inferring on CPU and up to 18\% on GPU, with less than 2\% drop in accuracy.

In the rest of the paper, we discuss our motivations in section \ref{sec:motivation}, the background and related works in section \ref{sec:bkg}, details of the method in section \ref{sec:method}, design and evaluation of the study in section \ref{sec:empirical-evaluation}, and lastly, we conclude the discussions in section \ref{sec:conclusion}.

\section{Motivation}\label{sec:motivation}
Many real-world software services utilize deep neural models and, simultaneously, require low response time to meet their service level objectives (SLO). This requirement usually leads to allocating expensive infrastructure and hardware resources to the services \citep{VelascoMontero2019OnTC}. The high computational cost of DNN models directly affects the service provider in terms of their delivery cost and the environment in terms of the carbon footprint of the data centers running such services 24/7. 

Countless high-traffic online platforms such as online stores, photo/video sharing platforms, digital advertising platforms, and trading platforms use neural networks within the process of serving their user requests. For instance, displaying an online advertisement involves an online ad-click rate prediction based on the user features \citep{Gharibshah2020DeepLF}. Furthermore, online stores also use deep learning classification models for various purposes, such as product categorization, recommendation, product review sentiment analysis, and customer churn rate prediction.

In terms of the traffic load, Google Lens for instance has reached an average of 3 billion usages per month in 2021 \citep{maxham_diaz_2021}. Employing a variety of machine learning and deep learning models, Google spends billions of dollars on data centers and infrastructure to process such volume of requests \citep{spadafora_2022}. Thus, extensive work towards minimizing the energy footprint of the large-scale services has been done \citep{Lo2014TowardsEP,Buyya2018SustainableCC}.

On the other hand, the trend in DNNs deployment on resource-constrained devices such as mobile and IoT devices has also been rising in the past few years \citep{Lin2020MCUNetTD,Yoo2020DeepLP}. Various scenarios involve DNNs performing on-device predictions where low inference latency and/or low compute consumption is required. For instance, traffic sign classification in autonomous vehicles \citep{Zhang2020LightweightDN} requires low latency, and on-device voice command recognition systems \citep{Lin2018EdgeSpeechNetsHE} and mobile visual assistants \citep{9179386} require low compute consumption.

Moreover, using pre-trained off-the-shelf DNN models and adapting them to new tasks using transfer learning is playing a fundamental role in enabling practitioners in different areas to utilize DNNs \citep{Shrestha2019CrossFrequencyCO,Abed2020AlzheimersDP,Lee2020EvaluationOT}. However, the pre-trained models' original training data is not always available to the users. The absence of the training data can be due to different reasons, such as the high volume or cost of the data, privacy requirements, or intellectual property regulations. Accounting for such common cases, we restrict our method to use only the data collected at inference time (test set). The inference data are unlabelled, meaning that their ground truth labels are not available to the user. Hence, our method relies only on the model's internal values and final outputs and does not require access to the ground truth labels.

DNNs compute performance improvement has received a considerable amount of attention in terms of specialized hardware accelerators \citep{Wang2019BenchmarkingTG,Dally2020DomainspecificHA,Deng2020ModelCA}, and framework-level optimizations \citep{Crankshaw2017ClipperAL,Shi2018PerformanceMA}. On the other hand, model compression methods propose modifications to the model's structure (i.e., weights and connections) to reduce their compute complexity. 

By applying one or more model compression methods, practitioners either replace the modified model and lose a fixed amount of accuracy, or manage multiple versions of the model with different accuracy and complexity. Having multiple model versions, they select one for inference based on the current workload \citep{Taylor2018AdaptiveDL, Marco2020OptimizingDL} or available resources \citep{Guan2018EnergyefficientAI}. However, our method optimizes the model while preserving its original structure, allowing the user to enable/disable the optimization without the overhead of managing and loading/offloading multiple model versions.

Considering the trends, requirements, and motivations discussed above, we design the caching method to add one or more alternative exit paths in the model with less computation required than the remaining layers in the backbone, controlled by the shallow classifiers we train using only the inference data. 

\section{Background and related works}\label{sec:bkg}
\begin{figure}[htbp]
\centering
\begin{center}
\includegraphics[width=\columnwidth]{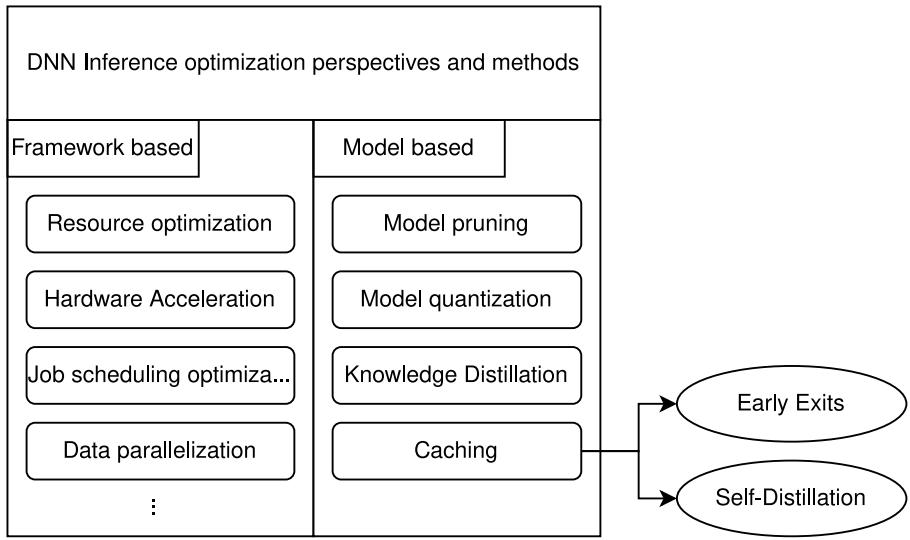}
\caption{DNN inference optimization perspectives and solutions}\label{fig:background}
\end{center}
\end{figure}

In this section, we briefly review the background topics to the model inference optimization problem. Following this background discussions, we introduce the techniques used to build the caching procedure.
Figure\ref{fig:background} puts the discussed background and related techniques into the picture.

\subsection{Inference optimization}
There are two perspectives addressing the model inference optimization problem.
The first perspective is interested in optimizing the model deployment platform and covers a broad range of optimization targets \citep{Yu2021ASO}. These studies often target the deployment environments in resource-constrained edge devices \citep{Liu2021SecDeepSA,Zhao2018DeepThingsDA} or resourceful cloud-based devices \citep{Li2020AutomatingCD}. Others focus on hardware-specific optimizations \citep{Zhu2018ResearchOP} and inference job scheduling \citep{Wu2020IrinaAD}.

The second perspective is focused on minimizing the model's inference compute requirements by compressing the model.
Among model compression techniques, model pruning \citep{han2015deep,Zhang2018ASD,Liu2019RethinkingTV}, model quantization \citep{Courbariaux2015BinaryConnectTD,Rastegari2016XNORNetIC,Nagel2019DataFreeQT}, and model distillation \citep{Bucila2006ModelC,Polino2018ModelCV,Hinton2015DistillingTK} are being extensively used. These ideas alleviate the model's computational complexity by pruning the weights, computing the floating-point calculations at lower precision, and distilling the knowledge from a teacher (more complex) model into a student (less complex) model, respectively. These techniques modify the original model and often cause a fixed amount of loss in the test accuracy.

\subsection{Early Exits in DNNs}\label{subsec:early-exit}
``Early exit'' generally refers to an alternative path in a DNN model which can be taken by a sample instead of proceeding to the next layers in the model. Many previous works have used the early exit concept for different purposes \citep{Xiao2021SelfCheckingDN,Scardapane2020WhySW,Matsubara2022SplitCA}.
Among them, Shallow Deep Networks (SDN) \citep{Kaya2019ShallowDeepNU} points out the ``overthinking'' problem in deep neural networks. ``Overthinking'' refers to the models spending a fixed amount of computational resources for any query sample, regardless of their complexity (i.e., how deep the neural network should be to infer the correct prediction for the sample). Their research proposes attaching shallow classifiers to the intermediate layers in the model to form the early exits. Each shallow classifier in SDN provides a prediction based on the values of the intermediate layer to which it is attached. 

On the other hand, \citep{Xiao2021SelfCheckingDN} incorporates the shallow classifiers to obtain multiple predictions for each sample. In their method, they use early exits as an ensemble of models to increase the base model's accuracy.

The functionality of the shallow classifiers in our proposed method is similar to SDN. However, the SDN method trains the shallow classifier using the ground truth data in the training set and overlooks the available knowledge in the original model. This constraint renders the proposed method useless when using a pre-trained model without access to the original training data, which is commonly the case for practitioners.

\subsection{DNN Distillation and Self-distillation}\label{subsec:distillation}
Among machine learning tasks, the classification category is one of the significant use cases where DNNs have been successful in recent years. Classification is applied to a broad range of data such as image \citep{Bharadi2017ImageCU}, text \citep{Varghese2020DeepLI}, audio \citep{Lee2009UnsupervisedFL}, and time-series \citep{Zheng2014TimeSC} classification.

Knowledge distillation(KD) \citep{Bucila2006ModelC,Polino2018ModelCV,Hinton2015DistillingTK} is a model compression method that trains a relatively small (less complex) model known as the student to mimic the behavior of a larger (more complex) model known as the teacher. Classification models usually provide a probability distribution (PD) representing the probability of the input belonging to each class. KD trains the student model to provide similar PDs (i.e.,\ soft labels) to the teacher model rather than training it with just a class label for each sample (i.e.,\ hard labels). KD uses specialized loss functions in the training process, such as Kullback-Leibler Divergence \citep{Joyce2011} to measure how one PD is different from another.

KD usually is a 2-step process consisting of training a large complex model to achieve high accuracy and distilling its knowledge into a smaller model. An essential challenge in KD is choosing the right teacher and student models. Self-distillation \citep{self-distillation} addresses this challenge by introducing a single-step method to train the teacher model along with multiple shallow classifiers. Each shallow classifier in self-distillation is a candidate student model which is trained by distilling the knowledge from one or more of the deeper classifiers. In contrast to SDN, self-distillation utilizes knowledge distillation to train the shallow classifiers. However, it still trains the base model from scratch along with the shallow classifiers, using the original training set. This training procedure conflicts with our objectives in both aspects. Specifically, we use a pre-trained model and keep it unchanged throughout the experiment and only use inference data to train the shallow classifiers.

Our work modifies and puts the presented methods in SDN and self-distillation in the context of caching the final predictions of pre-trained DNN models. The method trains the shallow classifiers using only the unlabelled samples collected at run-time and measures the improvement in inference compute costs achieved by the early exits throughout the forward-passes.

\subsection{DNN Prediction Caching}
Clipper \citep{Crankshaw2017ClipperAL} is a serving framework that incorporates caching DNNs predictions based on their inputs. Freeze Inference \citep{Kumar2019AcceleratingDL} investigates the use of traditional ML models such as K-NN and K-Means to predict based on intermediate layers' values. They show that the size and computation complexity of those ML models grows proportionally with the number of available samples and their computational overheads by far exceed any improvement. In Learned Caches, \citep{Balasubramanian2021AcceleratingDL} extend the Freeze Inference by replacing the ML models with a pair of DNN models. A predictor model predicting the outputs and a binary classifier predicting whether the output should be used as the final prediction. Their method uses the ground truth data in the process of training the predictor and selector models. In contrast, our method 1) only uses unlabelled inference data, 2) automates the process of cache-enabling, 3) uses a confidence-based cache hit determination, 4) handles batch processing by batch shrinking.

\section{Methodology}\label{sec:method}
In this section, we explain the method to convert a pre-trained deep neural model (which we call the backbone) to its extended version with our caching method (called cache-enabled model). The caching method adds one or more early-exit paths to the backbone, controlled by the shallow classifiers (which we call the cache models), allowing the model to infer a decision faster at run-time for some test data samples (cache hits). Faster decision for a portion of queries will result in a reduced mean response time.

``Cache model'' is a supplementary model that we attach to an intermediate layer in the backbone, and given the layer's values provides a prediction (along with a confidence value) for the backbone's output.
Just a reminder that as our principal motivation, we assume that the original training data is unavailable for the user, as is the case for most large-scale pre-trained models used in practice. Therefore, in the rest of the paper, unless we explicitly mention it, the terms dataset, training set, validation set, and test set all refer to the whole available data at run-time or a respective subset.


Our procedure for cache-enabling a pre-trained model is chiefly derived from the self-distillation method \citep{self-distillation}. However, we adopt the method to cache-enable pre-trained models using only their recorded outputs, without access to the ground truth (GT) labels.

A step-by-step guide on cache-enabling an off-the-shelf pre-trained model from a user perspective contains the following steps:
\begin{enumerate}
    \item Identify the candidate layers to be cached
    \item Build a cache model for each candidate
    \item Assign confidence thresholds to the built models for determining the cache hits
    \item Evaluate and optimize the cache-enabled model
    \item Update and maintenance
\end{enumerate}

In the following subsections, we further discuss the procedure and design decisions in each step outlined above.

\begin{figure}[ht]
\centering
\includegraphics[width=\columnwidth]{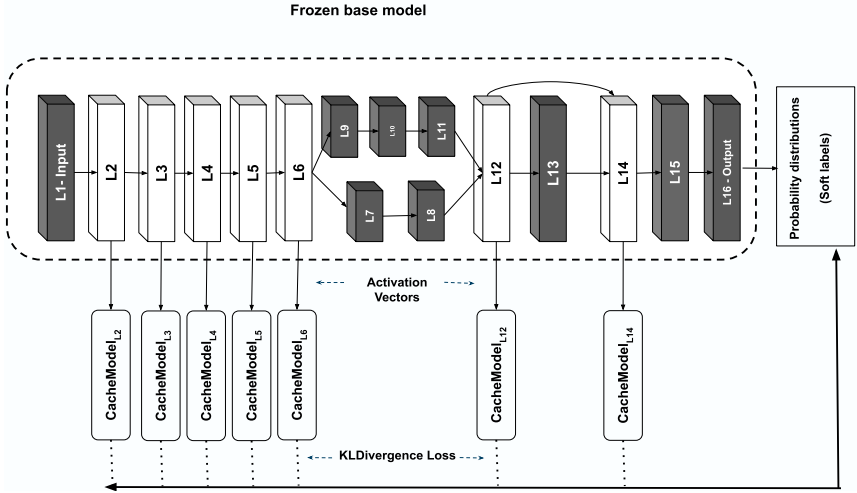}
\caption{Cache-enabling procedure, candidate layers, and data paths. }\label{fig:procedure}
\end{figure}

\subsection{Identifying candidate layers}\label{subsec:candidates}
Choosing which layers to cache is the first step toward cache-enabling a model. A candidate layer is a layer that we will examine its values correlation to the final predictions by training a cache model based on them. One can simply list all the layers in the backbone as candidates. However, since we launch a search for a cache model per candidate layer in the next step, we suggest narrowing the list by filtering out some layers with the following criteria:

\begin{itemize}
\item Some layers are disabled at inference time, such as dropouts and batch normalizations. These layers do not modify their input values at inference time. Therefore, we cross them off the candidates list.
\item A few last layers in the model (close to the output layer, such as \texttt{L15} in Figure \ref{fig:procedure}) might not be valuable candidates for caching since the remaining layers might not have heavy computations to reach the output. 
\item DNN models usually are composed of multiple components (i.e.\ first-level modules) consisting of multiple layers such as multiple residual blocks in ResNet models \cite{He2016DeepRL}). We narrow down the search space to the outputs layers in those components.
\item We only consider the layers which, given their activation values, the backbone's output is uniquely determined without any other layer's state involved (i.e., the backbone's output is a function of the layer's output). In other words, a layer with other layers or connections in parallel (such as \texttt{L7-L11} and \texttt{L13} in the Figure \ref{fig:procedure}) is not suitable for caching since the backbone's output does not solely depend on the layer's output.
\end{itemize}

Having the initial set of the candidate layers, we next build and associate a cache model to each one.

\subsection{Building cache models}\label{subsec:cache-models}
Building a cache model to be associated with an intermediate layer in the backbone consists of finding a suitable architecture for the cache model and training the model with that architecture. The details of the architecture search (search space, search method, and evaluation method) and the training procedure (training data extraction and the loss function) are discussed in the following two subsections.

\subsubsection{Cache models architecture}
A cache model can have an architecture with any size in depth and breadth, as long as it provides more computational improvement than its overhead. In other words, it must have substantially less complexity (i.e., number of parameters and connections) than the rest of the layers in the backbone that come after the corresponding intermediate layer. The search space for such models would contain architectures with different numbers and types of layers (e.g.,\ a stack of dense and/or convolution layers). Nevertheless, all the models in the search space must output a PD identical to the backbone's output in terms of size (i.e., the number of classes) and activation (e.g.,\ SoftMax or LogSoftMax). 

In our experiments, the search space consists of architectures with a stack of (up to 2) convolution layers followed by another stack of (up to 2) linear layers, with multiple choices of kernel and stride sizes for the convolutions and neuron counts for the linear layers. However, users can modify or expand the search space according to their specific needs and budget.

The objective of the search is to find a minimal architecture that converges and predicts the backbone's output with acceptable accuracy. Note that any accuracy given by a cache model (better than random) can be helpful as we will have a proper selection mechanism later in the process to only use the cache predictions that are (most likely) correct, and also to discard the cache models yielding low computational improvement.

The user can conduct the search by empirically sampling through the search space or by using a automated Neural Architecture Search (NAS) tool such as Auto-Keras \citep{Jin2019AutoKerasAE}, Auto-PyTorch \citep{zimmer2021auto}, Neural Network Intelligence (NNI) \citep{nni2021}, or NASLib \citep{naslib-2020}. However, we used NNI to conduct the search and customized the evaluation process to account for the models' accuracy and their computational complexity. We have used the floating point operations (FLOPs) count as the estimation for the models' computational complexity in this stage. 

Several factors influence a cache model's architecture for a given intermediate layer. These factors include the target intermediate layer's dimensions, its position in the backbone, and the dataset specifications such as its number of target classes. For instance, the first cache models in CIFAR100-Resnet50 and in CIFAR10-Resnet18 experiments (shown as cache1 in Figure \ref{fig:extended-schemas}) have the same input size, but since CIFAR100 has more target classes, it reasonably requires a cache model with more learning capacity. Therefore, using NAS to design the cache models helps automate the process and alleviate deep learning expert supervision in designing the cache models. 

Regardless of the search method, evaluating a nominated architecture requires training a model with the given architecture which we discuss the procedure in the next section. Moreover, since the search space is limited in depth, it is possible that for some intermediate layers, neither of the cache models converge (i.e., the model provides nearly random results). In such cases, we discard the candidate layer as non-suitable for caching.

\subsubsection{Training a cache model} \label{subsec:training}
Figure (\ref{fig:procedure}) illustrates a cache-enabled model's schema consisting of the backbone (the dashed box) and the associated cache models. A cache model's objective is to predict the output of the backbone model, given the corresponding intermediate layer's output, per input sample. 

Similar to the backbone, cache models are classification models. However, their inputs are the activation values in the intermediate layers. As suggested in self-distillation \citep{self-distillation}, training a cache model is essentially similar to distilling the knowledge from the backbone (final classifier) into the cache model.

Therefore, to distill the knowledge from the backbone into the cache models, we need a medial dataset (MD) based on the collected inference data (ID). The medial dataest for training a cache model associated to an intermediate layer \texttt{L} in the backbone \texttt{B} consists of the set of activation values in the layer \texttt{L} and the PDs given by \texttt{B} per samples in the given ID, formally annotated as below:

\begin{equation}
\label{eqn:InputLabelPairs1}
MD_L = [i \in ID : <B_L(i) ,  B(i)>]
\end{equation}
where:

\noindent
\details{\texttt{MD\textsubscript{L}}}{Medial dataset for the cache model associated with the layer \texttt{L}}
\details{\texttt{ID}}{The collected inference data consisting of unlabelled samples}
\details{\texttt{B\textsubscript{L}(i)}}{Activation values in layer \texttt{L} given the sample \texttt{i} to the backbone \texttt{B}}
\details{\texttt{B(i)}}{The backbone's PD output for the sample \texttt{i}}

Note that the labels in MDs are the backbone's outputs and not the GT labels, as we assumed the GT labels to be unavailable. We split the MD\textsubscript{L} into three splits ($MD_L^{Train}$, $MD_L^{Val}$, $MD_L^{Test}$) and use them respectively similar to the common deep learning training and test practices.

Similar to distillation method \citep{Hinton2015DistillingTK}, we use Kullback–Leibler Divergence (KLDiv) \citep{Joyce2011} loss function in the training procedure. KLDiv measures how different are the two given PDs. Thus, minimizing the KLDiv loss value over $MD_L^{Train}$ trains the cache model to estimate the prediction of the backbone ($B(i)$).

Unlike self-distillation where \citep{self-distillation} train the backbone and the shallow classifiers simultaneously, in our method, while training a cache model, it is crucial to freeze the rest of the model including the backbone and the other cache models (if any) in the collection, to ensure the training process does not modify any parameter not belonging to the current cache model.

\subsection{Assigning confidence threshold}
The probability value associated to the predicted class (the one with the highest probability) is known as the model's confidence in the prediction. The cache model's prediction confidence for a particular input will indicate whether we stick with that prediction (cache hit) or we proceed with the rest of the backbone to the next --- or probably final --- exit (cache miss).

Confidence calibration means enhancing the model to provide an accurate confidence. In other words, a well-calibrated model's confidence accurately represents the likelihood for that prediction to be correct(\cite{pmlr-v70-guo17a}). An over-confident cache model will lead the model to prematurely exit for some samples based on incorrect predictions, while an under-confident cache model will bear a low cache hit rate. Therefore, after building a cache model, we also calibrate its confidence using $MD_L^{Val}$ to better distinguish the predictions more likely to be correct. Several confidence calibration methods are discussed in \citep{pmlr-v70-guo17a}, among which the temperature scaling (in the output layer) has shown to be practical and easy to implement.

Having the model calibrated, we next assign a confidence threshold value to the model which will be used at inference time to determine the cache hits and misses. When a cache model identifies a cache hit, its prediction is considered to be the final prediction. However, when needed for validation and test purposes, we obtain the predictions from the cache model and the backbone.

\begin{table}[h]
\begin{center}
\begin{minipage}{\columnwidth}
\centering
\caption{Cache prediction confusion matrix, C: Cached predicted class, B: Backbone's predicted class, GT: Ground Truth label}\label{table:confusion}%
\begin{tabular}{@{}c|ccc@{}}
\toprule
Category & B = C & B = GT  &  C = GT  \\
\midrule
\textbf{$BC$}    & \checkmark   & \checkmark  & \checkmark  \\
$\overline{BC}$    & \checkmark & X   & X    \\
$B\overline{C}$   & X & \checkmark   & X   \\
$\overline{B}C$    & X & X   & \checkmark   \\
$\overline{B}\ \overline{C}$    & X & X   & X   \\
\end{tabular}
\end{minipage}
\end{center}
\end{table}

A cache model's prediction (C) for an input to the backbone falls into one of the 5 correctness categories listed in table \ref{table:confusion} with respect to the ground truth labels (GT) and the backbone's prediction (B) for the input. 

Among the cases where the cache model and the backbone disagree, the $B\overline{C}$ predictions negatively affect the final accuracy and on the other hand, the $\overline{B}C$ predictions positively affect the final accuracy. The Equation \ref{eqn:accuracy_change} formulates a cache model's actual effect on the final accuracy.

\begin{equation}
\label{eqn:accuracy_change}
F_\Delta(\theta) = \overline{B}C_\Delta(\theta) - B\overline{C}_\Delta(\theta) 
\end{equation}
Where:

\noindent
\details{$\Delta$}{The cache model}
\details{$F_\Delta$}{The actual accuracy effect $\Delta$ causes given $\theta$ as threshold}
\details{$B\overline{C}_\Delta$}{Ratio of $B\overline{C}$ predictions by $\Delta$ given $\theta$ as threshold}
\details{$\overline{B}C_\Delta$}{Ratio of $\overline{B}C$ predictions by $\Delta$ given $\theta$ as threshold}

However, since we use the unlabelled inference data to form the MDs, we can only estimate an upper bound for the cache model's effect in the final accuracy. The estimation assumes that an incorrect cache would always lead to an incorrect classification for the sample ($\overline{B}C$). We estimate the change in the accuracy upper bound a cache model causes given a certain confidence threshold, by its hit rate and cache accuracy:

\begin{equation}
\label{eqn:accuracy_change_ub}
F_\Delta(\theta) \le HR_\Delta(\theta) \times (1- CA_\Delta(\theta))
\end{equation}
Where 

\noindent
\details{$\Delta$}{The cache model}
\details{$F_\Delta$}{The expected accuracy drop $\Delta$ causes given $\theta$ as threshold}
\details{$HR_\Delta$}{Hit rate provided by $\Delta$ given $\theta$ as threshold}
\details{$CA_\Delta$}{Cache accuracy provided by $\Delta$ given $\theta$ as threshold}

Given the tolerance $T$ for drop in final accuracy, we assign a confidence threshold to each cache model that yields no more than $X/2^n\%$ expected accuracy drop on $MD_L^{Val}$ according to the Equation \ref{eqn:accuracy_change_ub}, where \texttt{n} is the 1-based index of the cache model in the setup.

It is important to note that there are alternative methods to distribute the accuracy drop budget among the cache models. For instance, one can equally distribute the budget. However, as we show in the evaluations later in section \ref{subsec:rq1}, we find it reasonable to assign more budget to the cache models shallower positions in the backbone.

\subsection{Evaluation and optimization of the cache-enabled model}\label{subsec:positions}

So far, we have a set of cached layers and their corresponding cache models ready for deployment. Algorithm \ref{alg:prediction} demonstrates a Python-style pseudo implementation of cache-enabled model inference process. When the cache-enabled model receives a batch of samples, it proceeds layer-by-layer similar to the standard forward-pass. Once a cached layer's activation values are available, it will pass the values to the corresponding cache model and obtains an early prediction with a confidence value per sample in the batch. For each sample, if the corresponding confidence value exceeds the specified threshold, we consider it a cache hit. Hence, we have the final prediction for the sample without passing it through the rest of the backbone. At this point, the prediction can be sent to the procedure awaiting the results (e.g.\ an API, a socket connection, a callback). We shrink the batch by discarding the cache hits items at each exit and proceed with a smaller batch to the next (or the final) exit.

\begin{algorithm}
    \caption{Cache-enabled model inference}\label{alg:prediction}
    \begin{algorithmic}[1]
    \Require Backbone \Comment{The original model}
    \Require CachedLayers \Comment{List of cached layers}
    \Require Layer \Comment{As part of Backbone, including the associated cache model and threshold}
    \Procedure{ForwardPass}{X, callback} \Comment{X: Input batch}
        \For{\texttt{Layer in Backbone.Layers}} \Comment{In order of presence}\footnotemark
            \State X $\gets$ Layer(X)
            \newline
            \If{Layer in CachedLayers}
                \State Cache $\gets$ Layer.CacheModel
                \State T $\gets$ Cache.Threshold
                \State cachedPDs $\gets$ Cache(X)
                \State confidences $\gets$ max(cachedPDs, axis=1)
                \State callback(cachedPDs[confidences$\geq$ T]) \Comment{Resolve cache hits}
                \State X $\gets$ X[confidences$<$T] \Comment{Shrink the batch}
            \EndIf
        \EndFor
    \EndProcedure
  \end{algorithmic}
\end{algorithm}
\footnotetext{The loop is to show that each cache model will receive the cached layer's activation values when available, immediately, before proceeding to the next layer in the base model.}

So far in the method, we have only evaluated the cache models individually, but to gain the highest improvement, we must also evaluate their collaborative performance within the cache-enabled model. Once the cache-enabled model is deployed, each cache model affects the following cache models' hit rates by narrowing the set of samples for which they will infer. More specifically, even if a cache model shows promising hit rate and accuracy in individual evaluation, its performance in the deployment can be affected due to the previous cache hits made by the earlier cache models (connected to shallower layers in the backbone). Therefore, we need to choose the optimum subset of cache models to infer the predictions with the minimum computations. 

A brute force approach to find the optimum subset would require evaluating the cache-enabled model with each subset of the cache models. However, we implement a more efficient method without multiple executions of the cache-enabled model.

First, for each cache model, we record its prediction per sample in the $MD_L^{Val}$ and their confidence values. 
We also record two FLOPs counts per cache model; One is the cache model's FLOPs count(C\textsubscript{1}), and the other is the fallback FLOPs count which denotes the FLOPs in the remaining layers in the backbone(C\textsubscript{2}). For example, for the layer \texttt{L12} in the Figure \ref{fig:procedure}, C\textsubscript{1} is the corresponding cache model's FLOPs count, and C\textsubscript{2} is the FLOPs count in the layers \texttt{L13} through \texttt{L16}.

For each subset $S$, we process the lists of predictions recorded for each model in $S$ to generate the lists of samples they actually receive when deployed along with other cache models in $S$. The processing consist of keeping only the samples in each list for which there has been no cache hits by the previous cache models in the subset. Further, we divide each list into two parts according to each cache model's confidence threshold; One consisting of the cache hits, and the other consisting of the cache misses.

Finally, we score each subset using the processed lists and recorded values for each cache model in $S$ as follows:

\begin{equation}
\label{eqn:cache_score}
K(S) = \sum_{\Delta\in S} \piped{H_\Delta} \times (C_{2, \Delta} - C_{1, \Delta}) - \piped{M_\Delta} \times C_{1, \Delta}
\end{equation}
Where

\noindent
\details{$K$}{The caching score for subset $S$}
\details{$\Delta$}{A cache model in $S$}
\details{$H_\Delta$}{The generated list of cache hits for $\Delta$}
\details{$M_\Delta$}{The generated list of cache misses for $\Delta$}
\details{$C_{1, \Delta}$}{FLOPs count recorded for $\Delta$}
\details{$C_{2, \Delta}$}{Fallback FLOPs count recorded for $\Delta$}

The score equation accounts for both the improvement a cache model provides through its cache hits within the subset, and the overhead it produces for its cache misses.

\begin{figure}[!htbp]
\centering
\includegraphics[width=\textwidth]{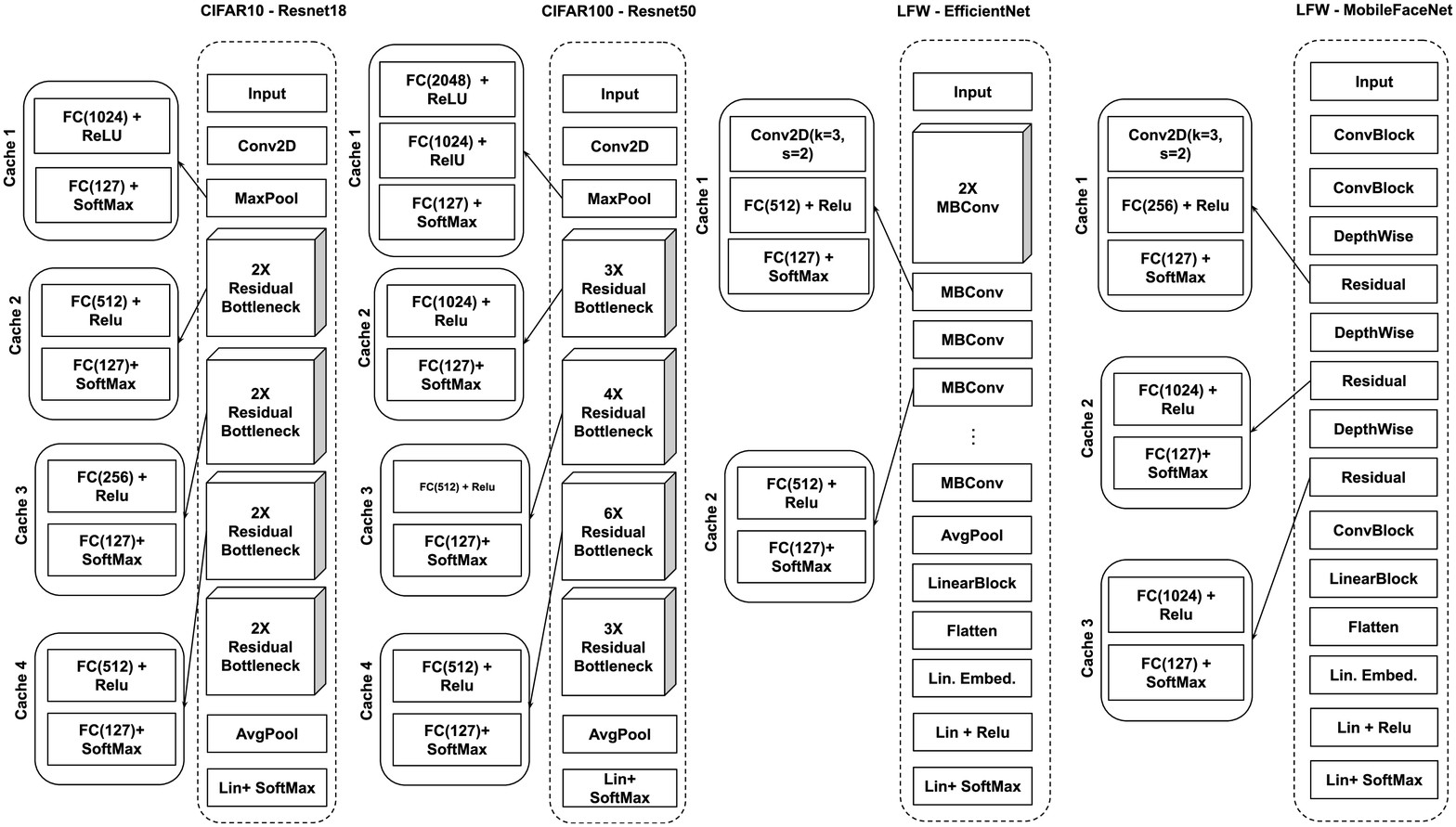}
\caption{Final schema of the cache models, for the experiments CIFAR10-Resnet18, CIFAR100-Resnet150, LWF-EfficientNet, and LFW-MobileFaceNet }\label{fig:extended-schemas}
\end{figure}

Final schemas after applying the method on MobileFaceNet, EfficientNet, ResNet18, and ResNet50 are illustrated in Figure \ref{fig:extended-schemas}. The figure demonstrates the chosen subsets and their associated cache models per backbone and dataset.

\subsection{Updates and maintenance}\label{subsec:maintenance}
Similar to conventional caching, layer caching also requires recurring updates to the cache space to adapt to the trend in inference data. However, unlike conventional caching, we can not update the cache models in real-time. Therefore, to update the cache models using the extended set of collected inference samples, we retrain them and re-adjust their confidence thresholds.

The retraining adapts the cache models to the trend in the incoming queries and maintains their cache accuracy. We consider two triggers for the updates: 
\begin{inparaenum}[I)]
\item When the size of the recently collected data reaches a threshold (e.g.\ 20\% of the collected samples are new) and
\item When the backbone is modified or retrained.
\end{inparaenum}
However, users must adapt the recommended triggers to their requirements and budget.


\section{Empirical Evaluation}\label{sec:empirical-evaluation}
In this section, we explain our experiment's objective, research questions, the tool implementation, and the experiment design including the backbones and datasets, evaluation metrics, and the environment configuration.

\subsection{Objectives and research questions}
The high-level objective of this experiment is to assess the ability of the automated layer caching mechanism to improve the compute requirements and inference time for DNN-based services. 

To address the above objective, we designed the following research questions (RQ):

\begin{itemize}
    \item [RQ1] To what extent the cache models can accurately predict the backbone's output and the ground truth data?\\
    This RQ investigates the core idea of caching as a mechanism to estimate the final outputs earlier in the model. The assessments in this RQ considers the cache models' accuracy in predicting the backbone's output (cache accuracy) and predicting the correct labels (GT accuracy).
    \item [RQ2] To what extent can cache-enabling improve compute requirements?\\
    In this RQ, we are interested in how cache-enabling affects the models' computation requirements. In these measurements, we measure the FLOPs counts and memory usage as the metrics for the models' compute consumption.
    \item [RQ3] How much acceleration does cache-enabling provide on CPU/GPU?\\
    In this RQ, we are interested in the actual amount of end-to-end speed up that a cache-enabled model can achieve. We break this result down to CPU and GPU accelerations, since they address different types of computation during the inference phase and thus may have been differently affected. 
    
\end{itemize}

\subsection{Tasks and datasets}\label{subsec:datasets}
Among the diverse set of classification tasks in real-world that are implemented by solutions utilizing DNN models, we have selected two representatives: face recognition and object classification. Both tasks are quite commonly addressed by DNNs and often used in large-scale services that have non-functional requirements such as: high throughput (due to the nature of the service and the large volume of input data) and are time-sensitive. 

The face recognition models are originally trained on larger datasets such as MS-Celeb-1M \citep{Guo2016MSCeleb1MAD} and are usually tested with different --- and smaller --- datasets such as LFW \citep{Huang2008LabeledFI}, CPLFW \citep{Zheng2017CrossAgeLA}, RFW \citep{Wang2019RacialFI}, AgeDB30 \citep{Moschoglou2017AgeDBTF}, and MegaFace \citep{KemelmacherShlizerman2016TheMB} for testing the models against specific challenges, such as age/ethnic biases, and recognizing mask covered faces.

We used the Labeled Faces in the Wild (LFW) dataset for face recognition which contains 13,233 images of 5,749 people. We used the images of 127 identities who have at least 11 images in the set so we can split them for training, validation and testing. 

We also used CIFAR10 and CIFAR100 test sets \citep{Krizhevsky2009LearningML} for object classification, each containing 10000 images distributed equally among 10 and 100 classes, respectively. 

A reminder that we do not use the training data, rather we only use the test sets to simulate incoming queries at run-time. Specifically, we use only the test splits of the CIFAR datasets. However, we use the whole LFW data as it has not been used to train the face recognition models. Moreover, we do not use the labels in these test sets in the training and optimization process, rather we only use them in the evaluation step to provide GT accuracy statistics.

Each dataset mentioned above represents an inference workload for the models. Thus, we split each one into training, validation and test partitions with 50\%, 20\%, and 30\% proportionality, respectively. However, we augmented the test sets using flips and rotations to improve the statistical significance of our testing measurements.

\subsection{Backbones}\label{subsec:backbones}
The proposed cache-enabling method is applicable to any deep classifier model. However, the results will vary for different models based on their complexity.

Among the available face recognition models, we have chosen well-known MobileFaceNet and EfficientNet models to evaluate the method, and we experiment with ResNet18 and ResNet50 for object classification.

The object classification models are typical classifier models out-of-the-box. However, the face recognition models are feature extractors that provide embedding vectors for each image based on the face/landmarks features. They can still be used to classify a face-identity dataset. Therefore, we attached a classifier block to those models and trained them (with the feature extractor layers frozen) to classify the images of the 127 identities with the highest number of images in the LFW dataset (above 10). It is important to note that since the added classifier block is a part of the pre-trained model under study, we discarded the data portion used to train the classifier block to ensure we still hold on to the constraint of working with pre-trained models without access to the original training dataset.

\subsection{Metrics and measurements}\label{sub:measurement}
Our evaluation metrics for RQ1 are ground truth (GT) accuracy and cache accuracy. 
Cache accuracy measures how accurately a cache model predicts the backbone's outputs (regardless of correctness). The GT accuracy applies to both cache-enabled model and each individual cache model. However, the cache accuracy only applies to the cache models.

In RQ2, we compare the original models and their cache-enabled version in terms of the average FLOPs count occurring for inference and their memory usage. We only measure the resources used in inference. Specifically, we exclude the training-specific layers (e.g., Batch Normalization and Dropout) and computations (e.g., gradient operations) in the analysis. 

FLOPs count takes the model architecture and the input size into account and estimates the computations required by the model to infer for the input \citep{Desislavov2021ComputeAE}. In other words, the fewer FLOPs used for inference, the more efficient is the model in terms of compute and energy consumption.

On the other hand, we report two aspects of memory usage for the models. First is the total space used to load the models on the memory (i.e.\ model size). This metric is essentially agnostic to the performance of cache models and only considers the memory cost of loading them along with the backbone. 

In addition to the memory required for their weights, DNNs also allocate a sizeable amount of temporary memory for buffers (also referred to as tensors) that correspond to intermediate results produced during the evaluation of the DNN's layers \cite{Levental2022MemoryPF}. Therefore, our second metric is the live tensor memory allocations (LTMA) during inference. LTMA measures the total memory allocated to load, move, and transform the input tensor through the model's layers to form the output tensor while executing the model.

In RQ3, we compare the average inference latency by the original model and its cache-enabled counterpart. Inference latency measures the time spent from passing the input to the model till it exits the model (by either an early exit or the final classifier in the backbone). Various factors affect the inference latency including hardware-specific optimizations (e.g., asynchronous computation), framework, and model implementation. In our measurements, the framework and model implementations are fixed as discussed in section \ref{subsec:implementation}. However, to account for other factors, we repeat each measurement for 100 times and report the average inference latency recorded for each experiment. Further, to also account for the asynchronous computations effects in GPU inference latency, we repeated the experiments with different batch sizes.
    

\subsection{Implementation}\label{subsec:implementation}
\begin{figure}[ht]
\centering
\includegraphics[width=\columnwidth]{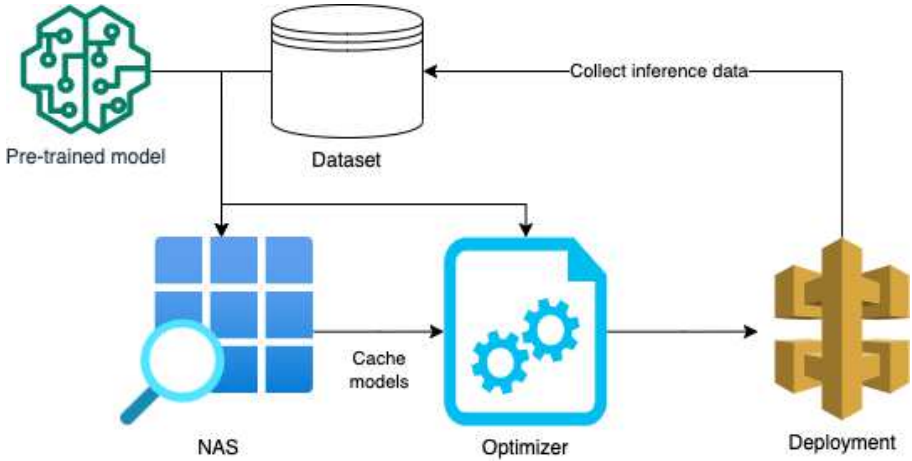}
\caption{Caching system overall framework}\label{fig:framework}
\end{figure}

We developed the caching tool using PyTorch, which is accessible through the GitHub repository\footnote{https://github.com/aminabedi/Automated-Layer-Caching}. Figure \ref{fig:framework} shows the overall system design. The tool provides a NAS module, an optimizer module, and a deployment module. The NAS module provides the architectures to be used per cache model. The optimizer assigns the confidence thresholds, finds the best subset of the cache models and provides evaluation reports. Lastly, the deployment module launches a web server with the cache-enabled model ready to serve queries.

\subsubsection{NAS Module}
Existing NAS tools typically define different search spaces according to different tasks which constrains their applicability to certain input types and sizes. Using such tools with input constraints defeats our method's generalization and automation purpose since the cache models' inputs can have any dimension and size. For instance, ProxylessNAS \citep{Cai2019ProxylessNASDN} specializes in optimizing the neural architecture performance for a target hardware. However, it is only applicable for image classification tasks and requires certain input specifications (e.g.,\ 3xHxW images normalized using given values). Similarly, Auto-PyTorch \citep{zimmer2021auto} and Auto-Keras are only applicable to tabular, text, and image datasets. 

We chose NNI by Microsoft \citep{nni2021} as it does not constrain the model inputs in terms of type, size, and dimensions. NNI also provides an extensible search space definition with support for variable number of layers and nested choices (e.g., choosing among different layer types, each with different layer-specific parameters).

Given the backbone implementation, the dataset, and the search space, the module launches an NNI experiment per candidate layer to search for an optimum cache model for the layer. Each experiment launches a web GUI for the progress reports and the results.

We aim for end-to-end automation in the tool. However, currently, the user still needs to manually export the architecture specifications when using the NAS module and convert them to a proper python implementation (i.e., a PyTorch module implementing the architecture). The specifications are available to the user through the experiments web GUI and also in the trial output files. This shortcoming is due to the NNI implementation, which does not currently provide access to the model objects within the experiments. We have created an enhancement suggestion on the NNI repository to support the model object access (issue \#4910).

\subsubsection{Optimizer and deployment modules}
Given the backbone's implementation and the cache models, the optimizer evaluates cache models, assigns their confidence thresholds, finds the best subset of the cache models and disables the rest, and finally reports the relevant performance metrics for the cache-enabled model and each cache model. We used the DeepSpeed by Microsoft and PyTorch profiler to profile the FLOPs counts, memory usage, and latency values for the cache models and the backbones.

The user can use each module independently. Specifically, the user can skip the architecture search via the NAS module and provide the architectures manually to the optimizer, and the module trains them before proceeding to the evaluation.

The tool also offers an extensive set of configurations. More specifically, the user can configure the tool to use one device (e.g., GPU) for training processes and the other (e.g., CPU) for evaluation and deployment.

The deployment module launches a web server and exposes a WebSocket API to the cache-enabled model. The query batches passed to the socket will receive one response per item, as soon as the prediction is available through either of the (early or final) exits.

\subsubsection{Backbone Implementation}
We used the backbone implementations and weights provided by the FaceX-Zoo \citep{Wang2021FaceXZooAP} repository to conduct the experiments with LWF dataset on MobileFaceNet and EfficientNet models.

For experimenting with CIFAR10 and CIFAR100, we used the implementations provided by torchvision \citep{Marcel2010TorchvisionTM} and the weights provided by \citep{huy_phan_2021_4431043} and \citep{weiaicunzai2020}.

All the backbone implementations were modified to implement an interface that handles the interactions with the cache models, controls the exits (cache hits and misses), and provides the relevant reports and metrics. We documented the interface usage in the repository, so users can experiment with new backbones and datasets. We refer interested readers to a blog post on how to extract intermediate activations in PyTorch \citep{nanbhas2020forwardhook} which introduces three methods to access the activation values. The introduced forward hooks method in PyTorch is very convenient for read-only purposes. However, our method requires performing actions based on the activation values, specifically, cache lookup and batch shrinking and avoiding further computation through the next layers. Therefore, we used the so called ``hacker'' method to access the activation values and perform these action and provided the interface for easy replication on different backbones.

\subsection{Environment setup}
The hardware used for inference substantially affects the results due to the hardware-specific optimizations such as computation parallelism. In our experiments, we have used an ``Intel(R) Core(TM) i7-10700K CPU @ 3.80GH'' to measure on-CPU inference times and an ``NVIDIA GeForce RTX 3070'' GPU to measure on-GPU inference time.

\subsection{Experiment results}\label{sec:evaluation}
In this sub-section, we evaluate the results of applying the method on the baseline backbones and discuss the answers to the RQs.

\subsubsection{RQ1. How accurate are the cache models in predicting the
backbone's output and the ground truth labels?}\label{subsec:rq1}

In this RQ we are interested in the built cache models' performance in terms of their hit rate, GT accuracy, and cache accuracy. We break down the measurements into two parts. The first part covers the cache models' individual performance over the whole test set without any other cache model involved. The second part covers their collaborative performance within in the cache-enabled model.

\subsubsection{Cache models' individual performance}\label{subsubsec:individual-performance}
\begin{figure}[!htbp]
\centering
\includegraphics[width=\columnwidth]{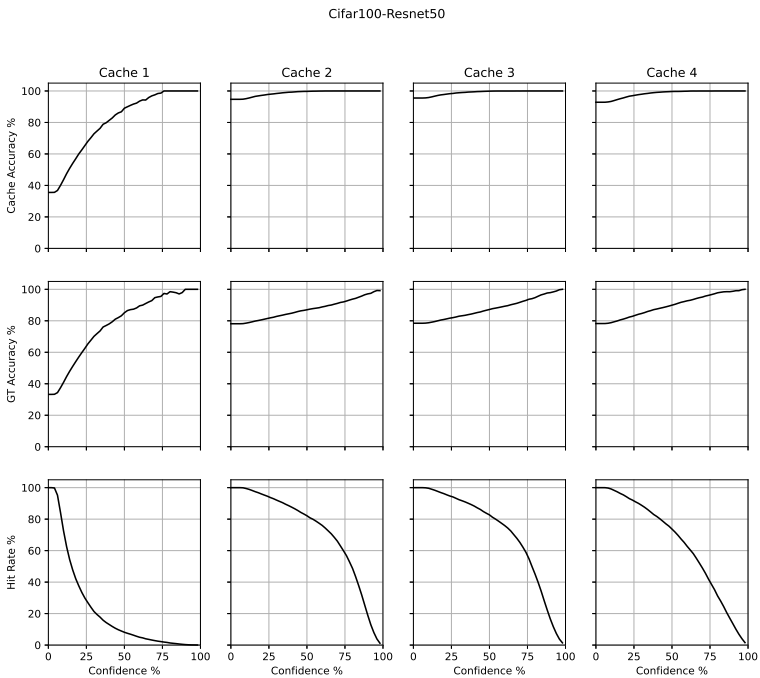}
\caption{Individual accuracy and hit rate of the cache models vs. confidence threshold per cache model in CIFAR100 - Resnet50 experiment}\label{fig:acc-conf-cifar100-resnet50}
\end{figure}
Figure \ref{fig:acc-conf-cifar100-resnet50} portrays each cache model's individual performance against any confidence threshold value in CIFAR100-Resnet50 experiment. The figures demonstrating the same measurements for other experiments are available in appendix \ref{appendice1}.

We make three key observations here. 
First, deeper cache models are more confident and accurate in their predictions. For instance, cache 1 in the Figure \ref{fig:acc-conf-cifar100-resnet50} has 33.36\% GT accuracy and 35.74\% cache accuracy, while these metrics increase to 78.60\% and 95.38\% for Cache 3, respectively. This observation agrees with the generally acknowledged feature extraction pattern in the DNNs --- deeper layers convey more detailed information.

The second key observation is the inverse correlation between the cache models' accuracy (both GT and cache) and their hit rates. This observation highlights the reliability of confidence thresholds in distinguishing the predictions more likely to be correct. For instance, cache 1 in Figure \ref{fig:acc-conf-cifar100-resnet50}, with a 20\% confidence threshold, yields 35.24\% hit rate but also 8.99\% drop in the final accuracy. However, with a 60\% confidence threshold, it yields a 4\% hit rate and does not reduce the final accuracy more than 0.1\%.

The third observation is that the cache accuracy is higher than the GT accuracy in all cases. This difference is because we have trained the cache models to mimic the backbone only by observing its activation values in the intermediate layers and outputs. Since we have not assumed access to the GT labels (which is the case for inference data collected at run-time) while training the cache models, they have learned to make correct predictions only through predicting the backbone's output, which might have been incorrect in the first place. On the other hand, we observed that the cache models predict the correct labels for a portion of samples for which the backbone misclassifies. For instance, for 0.92\% of the samples, cache 3 (in the Figure \ref{fig:acc-conf-cifar100-resnet50}) correctly predicted the GT labels while the backbone failed ($\overline{B}C$ predictions). This shows the cache models' potential to partially compensate for their incorrect caches ($B\overline{C}$ predictions) by correcting the backbone's predictions for some samples ($\overline{B}C$). This indeed agrees with the overthinking concept in SDN (as discussed in \ref{subsec:distillation}) since for this set of samples, the cache models have been able to predict correctly in the shallower layers of the backbone.

\subsubsection{Cache models' collaborative performance}
\begin{table}[!htbp]
\begin{center}
\begin{minipage}{\columnwidth}
\caption{Cache models' collaborative performance in terms of hit rate(HR), cache accuracy (A\textsubscript{cache}), GT accuracy (A\textsubscript{GT}), and their effect on the final accuracy($\downarrow$A\textsubscript{effect}). LFW: Labeled Faces in the Wild, MFN: MobileFaceNet, EFN: EfficientNet}\label{tab:collaboration}%
\begin{tabular}{@{}ccccc|cccc@{}}
\toprule
\multirow{2}{*}{Data}&\multirow{2}{*}{Model}&\multicolumn{2}{c}{Final accuracy}&\multirow{2}{*}{Exit\#}  & \multirow{2}{*}{HR} & \multirow{2}{*}{A\textsubscript{cache}} & \multirow{2}{*}{A\textsubscript{GT}} & \multirow{2}{*}{$\downarrow$ A\textsubscript{effect}}\\
& &  Base & Cache-enabled &  & & & \\
\midrule
 \multirow{8}{*}{\rotatebox[origin=c]{90}{CIFAR10}} & \multirow{4}{*}{\rotatebox[origin=c]{90}{Resnet18}} & \multirow{4}{*}{88.71\%} & \multirow{4}{*}{86.49\%} & 1 & 67.21\% & 92.29\% & 88.91\% & 01.31\% \\
 & & & & 2 & 10.33\% & 89.76\% & 76.63\% & 0.56\% \\
 & & & & 3 & 11.24\% & 85.71\% & 51.43\% & 0.25\%\\
 & & & & 4 & 8.32\% & 91.37\% & 35.71\% & 0.1 \%\\
 & \multirow{4}{*}{\rotatebox[origin=c]{90}{Resnet50}} & \multirow{4}{*}{87.92\%} & \multirow{4}{*}{85.88\%} & 1 & 61.41\% & 89.12\% & 86.19\% & 1.12\%\\
 & & & & 2 & 15.73\% & 93.01\% & 77.84\% & 0.58\%\\
 & & & & 3 & 10.29\% & 82.22\% & 53.33\% & 0.3\%\\
 & & & & 4 & 6.1\% & 97.47\% & 42.65\% & 0.04\%\\
 \hline
 \multirow{8}{*}{\rotatebox[origin=c]{90}{CIFAR100}} & \multirow{4}{*}{\rotatebox[origin=c]{90}{Resnet18}} & \multirow{4}{*}{75.92\%} & \multirow{4}{*}{74.47\%} & 1 & 11.96\% & 99.29\% & 82.11\% & 0.94\% \\
 & & & & 2 & 58.26\% & 99.62\% & 85.41\% & 0.1\% \\
 & & & & 3 & 7.26 \% & 93.81\% & 59.29\% & 0.3\%\\
 & & & & 4 & 5.36\% & 55.56\% & 38.89\% & 0.11\%\\
 & \multirow{4}{*}{\rotatebox[origin=c]{90}{Resnet50}} & \multirow{4}{*}{78.98\%} & \multirow{4}{*}{77.04\%} & 1 & 11.92\% & 76.34\% & 80.2\% & 1.32\%\\
 & & & & 2 & 61.98\% & 98.56\% & 84.55\% & 0.34\%\\
 & & & & 3 & 11.5\% & 97.85\% & 63.69\% & 0.27\%\\
 & & & & 4 & 7.38\% & 73.68\% & 52.63\% & 0.1\%\\
 \hline
 \multirow{5}{*}{\rotatebox[origin=c]{90}{LFW}} & \multirow{3}{*}{\rotatebox[origin=c]{90}{MFN}} & \multirow{3}{*}{97.78\%} & \multirow{3}{*}{96.91\%} & 1 & 37.35\% & 98.63\% & 97.88\% & 0.51\% \\
 & & & & 2 & 41.02\% & 99.71\% & 99.71\% & 0\% \\
 & & & & 3 & 55.95\% & 93.44\% & 96.18\% & 0.24\% \\

 & \multirow{2}{*}{\rotatebox[origin=c]{90}{EFN}} & \multirow{2}{*}{97.29\%} & \multirow{2}{*}{95.35\%} & 1 & 63.73\% & 96.82\% & 96.24\% & 1.67\%\\
 & & & & 2 & 14.52\% & 99.12\% & 98.76\% & 0.02\%\\
 \hline

\end{tabular}
\end{minipage}
\end{center}
\end{table}
Table \ref{tab:collaboration} describes the cache models' collaborative performance within the cache-enabled model per experiment. In the table, we also report how each cache model's cache hits have affected the final accuracy. 

Here, we observe that while evaluating the cache models on the subset of samples, which were missed by the previous cache models (the relatively more complex ones), the measured hit rate and GT accuracy is substantially lower compared to the evaluation on the whole dataset. This is indeed due to the fact that the simpler samples (less detailed and easier to classify) are resolved earlier in the model. More specifically, hit rate decreases since the cache models are less confident in their prediction for the more complex samples, and GT accuracy also decreases since the backbone also is less accurate for such samples.
However, we observe that the cache models still have high cache accuracy with low impact on the overall accuracy. This observation shows how the confidence-based caching method has effectively enabled the cache models to provide early predictions and keep the overall accuracy drop within the given tolerance.

\subsubsection{RQ2. To what extent can cache-enabling improve compute requirements?}\label{subsec:rq2}
In this RQ, we showcase the amount of computation caching can save in terms of FLOPs count and analyze the memory usage of the models.

\begin{table}[!htbp]
\begin{center}
\begin{minipage}{\columnwidth}
\centering
\caption{Original and cache-enabled models FLOPs (M:Mega - $10^6$)}\label{table:flops}
\begin{tabular}{@{}ll|ccc@{}}
\toprule
\multirow{2}{*}{Dataset(input size)} & \multirow{2}{*}{Model} &  \multicolumn{2}{c}{FLOPs} & \multirow{2}{*}{$\downarrow$ Ratio} \\
& & Original & Cache-enabled \\
\midrule
 \multirow{2}{*}{CIFAR10($3\times32\times32$)} & Resnet18 & 765M & 414M & 45.88\%\\
 & Resnet50 & 1303M & 601M & 53.87\%\\
 \hline
 \multirow{2}{*}{CIFAR100($3\times32\times32$)} & Resnet18 & 766M & 374M & 51.17\%\\
 & Resnet50  & 1304M & 547M & 58.05\%\\
 \hline
 \multirow{2}{*}{LFW($3\times112\times112$)} & MobileFaceNet  & 474M & 296M & 37.55\% \\
 & EfficientNet  & 272M & 182M & 33.08\% \\

\end{tabular}
\end{minipage}
\end{center}
\end{table}

Table \ref{table:flops} demonstrates the average amount of FLOPs computed for inference per sample. Here we observe that shrinking the batches proportionally decreases the FLOPs count required for inference.

\begin{table}[!htbp]
\begin{center}
\begin{minipage}{\columnwidth}
\centering
\caption{Original and cache-enabled models memory usage}\label{table:memory}
\begin{tabular}{@{}ll|cccc@{}}
\toprule
\multirow{3}{*}{Dataset(input size)} & \multirow{3}{*}{Model} & \multicolumn{2}{c}{Original} &  \multicolumn{2}{c}{Cache-enabled} \\
& & Model Size & LTMA & Model Size & LTMA \\
\midrule
 \multirow{2}{*}{CIFAR10($3\times32\times32$)} & Resnet18 & 43MB & 102MB & 97MB & 88MB\\
 & Resnet50 & 91MB & 235MB & 243MB & 201MB\\
 \hline
 \multirow{2}{*}{CIFAR100($3\times32\times32$)} & Resnet18 & 43MB & 104MB & 383MB & 93MB\\
 & Resnet50  & 91MB & 235MB & 552MB & 189MB\\
 \hline
 \multirow{2}{*}{LFW($3\times112\times112$)} & MobileFaceNet & 286MB & 567MB & 350MB & 515MB \\
 & EfficientNet & 147MB & 371MB & 297MB & 349MB \\

\end{tabular}
\end{minipage}
\end{center}
\end{table}

Moreover, table \ref{table:memory} shows the memory used to load the models (i.e.,\ the model size) and the total LTMA during inference while inferring for the test set.  As expected, the cache-enabled models' size is larger than the original model in all cases since they include the backbone and the additional cache models. However, the decreased LTMA in all cases shows the reduced amount of memory allocations during the inference time. Generally, lower LTMA indicates smaller tensor dimensions (e.g.\ batch size, input and operators' dimensions) \citep{Ren2021SentinelET}. However, in our case, since we do not change neither of the dimensions, lower LTMA is due to avoiding the computations in the remaining layers after cache hits which require further memory allocations. 

Although the FLOPs count and memory usage indicate the model's inference computational requirements, the decreased amount of FLOPs and LTMA does not necessarily lead to proportional reduction in the models' inference latency, which we further investigate in the next RQ.
\subsubsection{RQ3. How much acceleration does cache-enabling provide on CPU/GPU?}\label{subsec:rq3}
In this RQ, we investigate the end-to-end improvement that cache-enabling offers.
The results of this measurement clearly depend on multiple deployment factors such as the underlying hardware and framework, and as we discuss later in the section, their asynchronous computation capabilities.

\begin{table}[h]
\begin{center}
\begin{minipage}{\columnwidth}
\centering
\caption{end-to-end evaluation of cache-enabled models improvement in average inference latency, batch size = 32, MFN: MobileFaceNet, EFN: EfficientNet}\label{table:latencies}%
\begin{tabular}{@{}ll|cccccc@{}}
\toprule
\multirow{2}{*}{Dataset} &\multirow{2}{*}{Model} &  \multicolumn{2}{c}{Original latency} & \multicolumn{2}{c}{Cache-enabled latency} & \multicolumn{2}{c}{$\downarrow$ Ratio}\\
& &  CPU & GPU & CPU & GPU & CPU & GPU \\
\midrule
 \multirow{2}{*}{CIFAR10} &Resnet18 & 13.4 ms& 1.08 ms& 10.11 ms& 0.98 ms& 24.55\%& 10.2\% \\
 & Resnet50 & 18.73 ms & 1.81 ms & 14.62 ms & 1.51 ms & 31.08\% & 16.57\% \\
 \hline
 \multirow{2}{*}{CIFAR100} & Resnet18 & 14.23 ms& 1.39 ms& 9.39 ms& 1.25 ms& 34.01\%& 10.08\%\\
 & Resnet50  & 19.59 ms & 2.05 ms & 9.02 ms & 1.84 ms & \textbf{46.08\%} & 16.75\% \\
 \hline
 \multirow{2}{*}{LFW} & MFN  & 25.34 ms & 8.22 ms & 16.91 ms & 7.30 ms & 33.23\% & 11.19\%  \\
 & EFN  & 39.41 ms & 17.63 ms & 27.98 ms & 14.38 ms & 29.01\% & \textbf{18.44}\%  \\

\end{tabular}
\end{minipage}
\end{center}
\end{table}

Table (\ref{table:latencies}) shows the average latency for the base models on CPU and GPU, vs. their cache-enabled counterparts, evaluated on the test set. 

The first key observation here is the improvements on CPU. This improvement is due to the low parallelism in the CPU architecture. Essentially, the computations volume on CPU is proportional to the number of samples. Therefore, when a sample takes an early exit, the remaining computation required to finish the tasks for the batch proportionally decreases. 

The second observation is the relatively lower latency improvement on GPU. This observation shows that shrinking a batch does not proportionally reduce the inference time on GPU, which is due to the high parallelism in the hardware. Shrinking the batch on GPU provides a certain overhead since it interrupts the on-chip parallelism and hardware optimizations. This interruption forces the hardware to re-plan its computations which can be time consuming. Thus, batch shrinking improvements can be insignificant on GPU.

\begin{table}[h]
\begin{center}
\begin{minipage}{\columnwidth}
\centering
\caption{Inference latency improvement on GPU vs. batch size in Resnet18 and Resnet50 trained on CIFAR100}\label{table:gpu-batch}%
\begin{tabular}{@{}lc|cccccc@{}}
\toprule
Model & Batch Size & Original Latency & Cache-enabled Latency & $\downarrow$ Ratio\\
\midrule
\multirow{4}{*}{Resnet18} & 16 & 1.34 ms & 1.18 ms & 11.83\% \\
& 32 & 1.39 ms & 1.25 ms & 10.08\% \\
& 64 & 1.43 ms & 1.77 ms & -24.28\% \\
& 128 & 1.61 ms & 2.11 ms & -31.05\% \\
 \hline
\multirow{4}{*}{Resnet50} & 16 & 1.98 ms & 1.71 ms & 13.68\% \\
& 32 & 2.05 ms & 1.84 ms & 16.75\% \\
& 64 & 2.19 ms & 1.98 ms & 9.21\% \\
& 128 & 2.7 ms & 3.22 ms & -19.43\% \\
\end{tabular}
\end{minipage}
\end{center}
\end{table}

Table \ref{table:gpu-batch} further demonstrates how the batch size affects the improvement provided by caching. The key observation here is that increasing the batch size can negate the caching effect on the inference latency which as discussed is due to fewer number of batches that are fully resolved through the cache models and do not reach the last layers. In conclusion, the latency improvement here highly depends on the hardware used in inference and must be specifically analyzed per hardware environment and computation parameters such as batch size.
However, the method still can be useful when the model is not performing batch inferences (batch size = 1). One can also use the tool and get a best prediction so far within the forward-pass process by disabling the batch shrinking. Doing so will generate multiple predictions per input sample, one per exit (early and final).

\subsection{Limitation and future directions}\label{subsec:discussion}
The first limitation of this study is that the proposed method is limited to classification models since it would be more complicated for the cache models to predict a regression model's output due to their continuous values. This limitation is strongly tied to the effectiveness of knowledge distillation in case of regression models. 

The method also does not take the internal state of the backbone (if any) into account, such as the hidden states in recurrent neural networks. Therefore, the method's effectiveness for such models still needs to be further assessed.

Moreover, practitioners should take the underlying hardware and the backbone structure into account as they directly affect the final performance. On this note, as shown in section \ref{subsec:rq3}, different models provide different performances in terms of inference latency in the first place, therefore, choosing the right model for the task comes first, and caching can be helpful in improving the performance.

\section{Conclusion}\label{sec:conclusion}
In this paper, we have showed that our automated cashing approach is able to extend a pre-trained classification DNN to a cache-enabled version using a relatively small and unlabelled dataset. The required training dataset for cashing models are collected just by recording the input items and their corresponding backbone outputs at the inference time.
We have also shown that the caching method can introduce significant improvement in the model's computing requirements and inference latency, specially when the inference is performed on CPU.

We discussed the parameters, design choices, and the procedure of cache-enabling a pre-trained off-the-shelf model, and the required updates and maintenance.

In conclusion, while traditional caching might not be beneficial for DNN models due to the diversity, size and dimensions of the inputs, caching the features in the hidden layers of the DNNs using the cache models can achieve significant improvement in the model's inference computational complexity and latency. As shown in sections \ref{subsec:rq2} and \ref{subsec:rq3}, caching reduces the average inference FLOPs by up to 58\% and the latency up to 46.09\% on CPU and 18.44\% on GPU.

\backmatter





\bmhead{Acknowledgments}

The work of Pooyan Jamshidi has been partially supported by NSF (Awards 2007202, 2107463, and 2233873) and NASA (Award 80NSSC20K1720).


\section*{Declarations}

\textbf{Conflict of interest} There are no conflict of interests.







\begin{appendices}

\section{Cache models' individual performance for all experimenst}\label{appendice1}

The following figures demonstrate the hit rate, GT accuracy, and cache accuracy of each cache model vs. the confidence threshold, per experiment dataset and backbone.
\begin{figure}[p]
\centering
\includegraphics[width=\columnwidth]{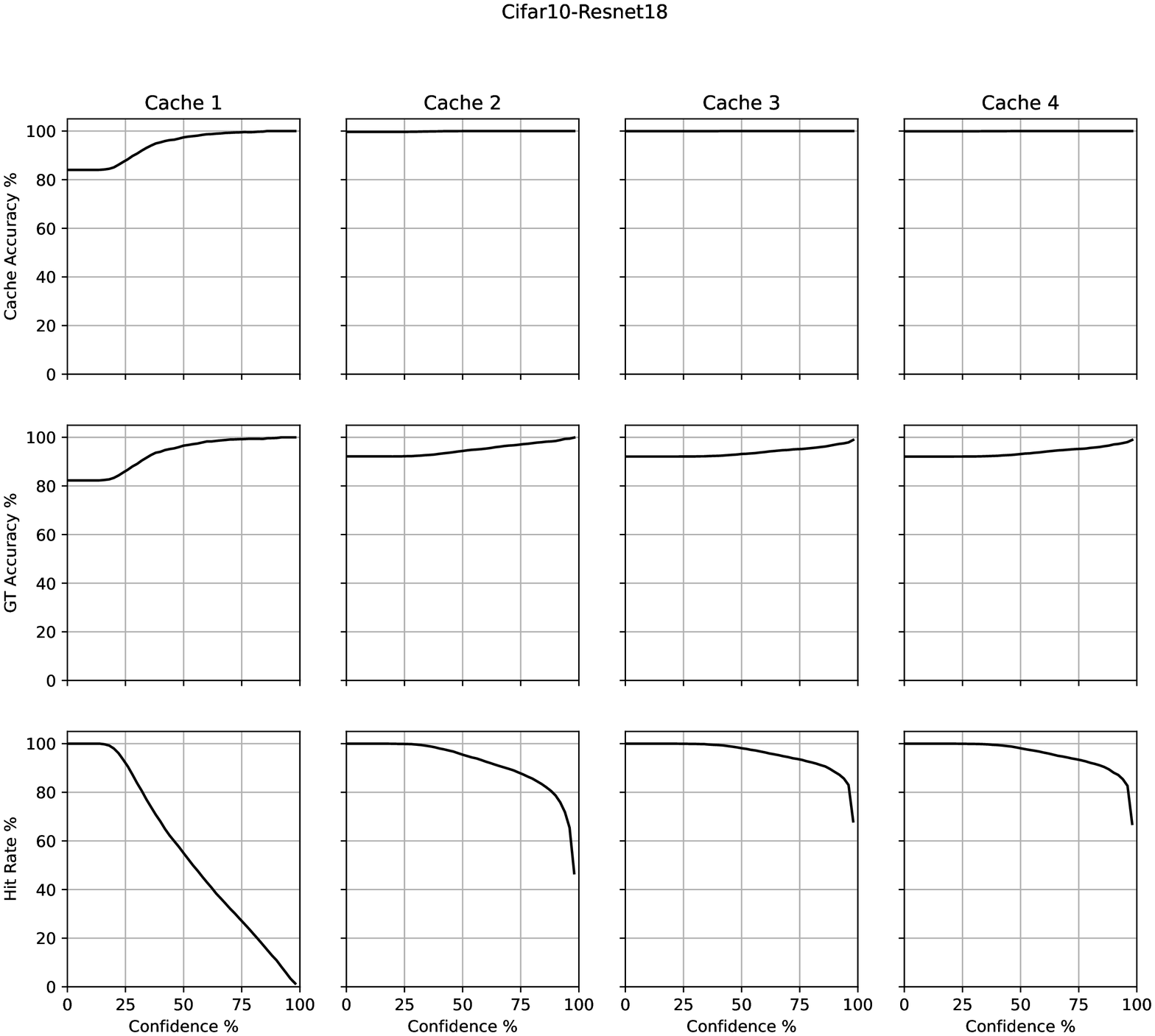}
\caption{Experiment: CIFAR10-Resnet18}
\end{figure}

\begin{figure}[p]
\centering
\includegraphics[width=\columnwidth]{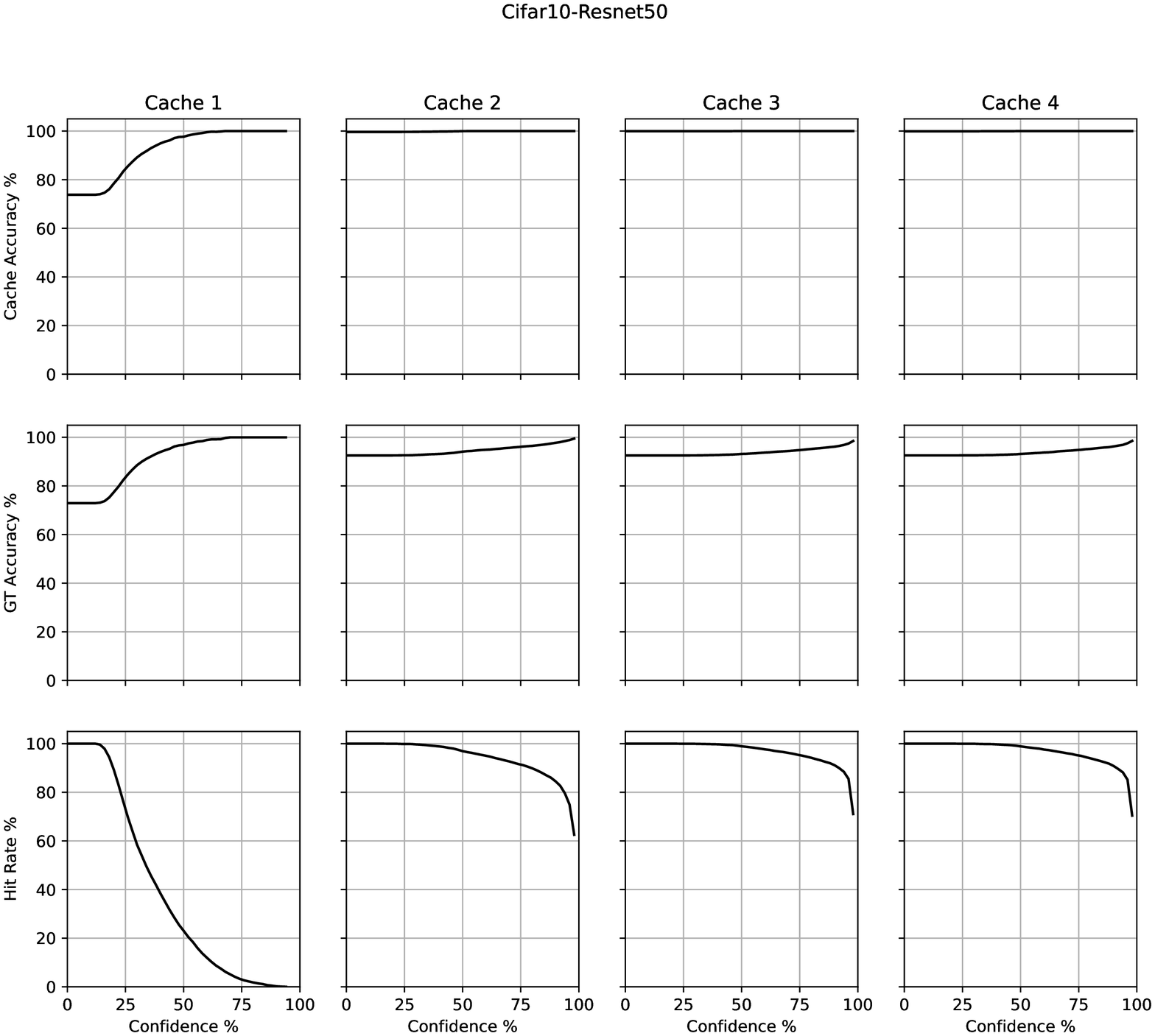}
\caption{Experiment: CIFAR10-Resnet50}
\end{figure}

\begin{figure}[p]
\centering
\includegraphics[width=\columnwidth]{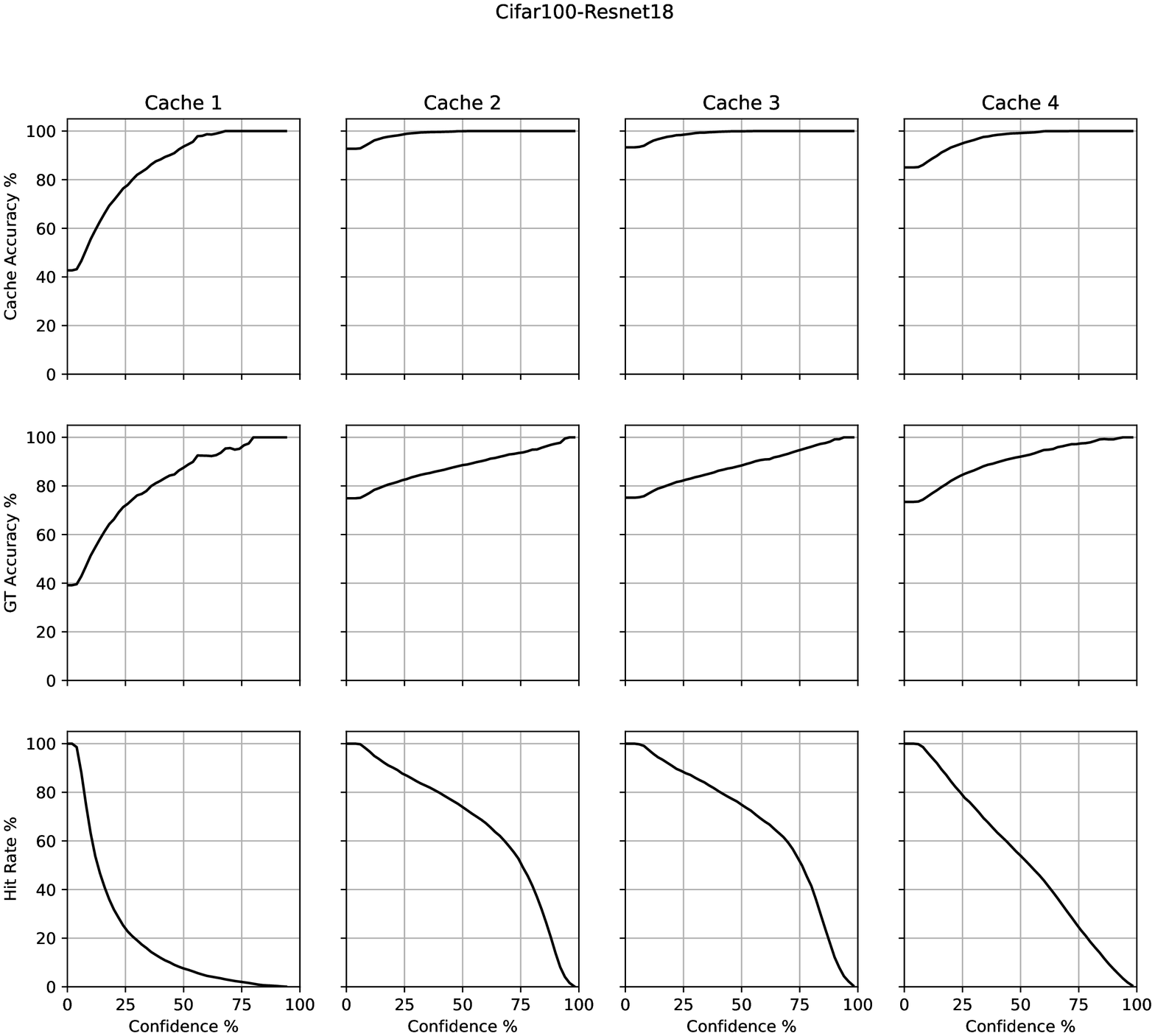}
\caption{Experiment: CIFAR100-Resnet18}
\end{figure}

\begin{figure}[p]
\centering
\includegraphics[width=\columnwidth]{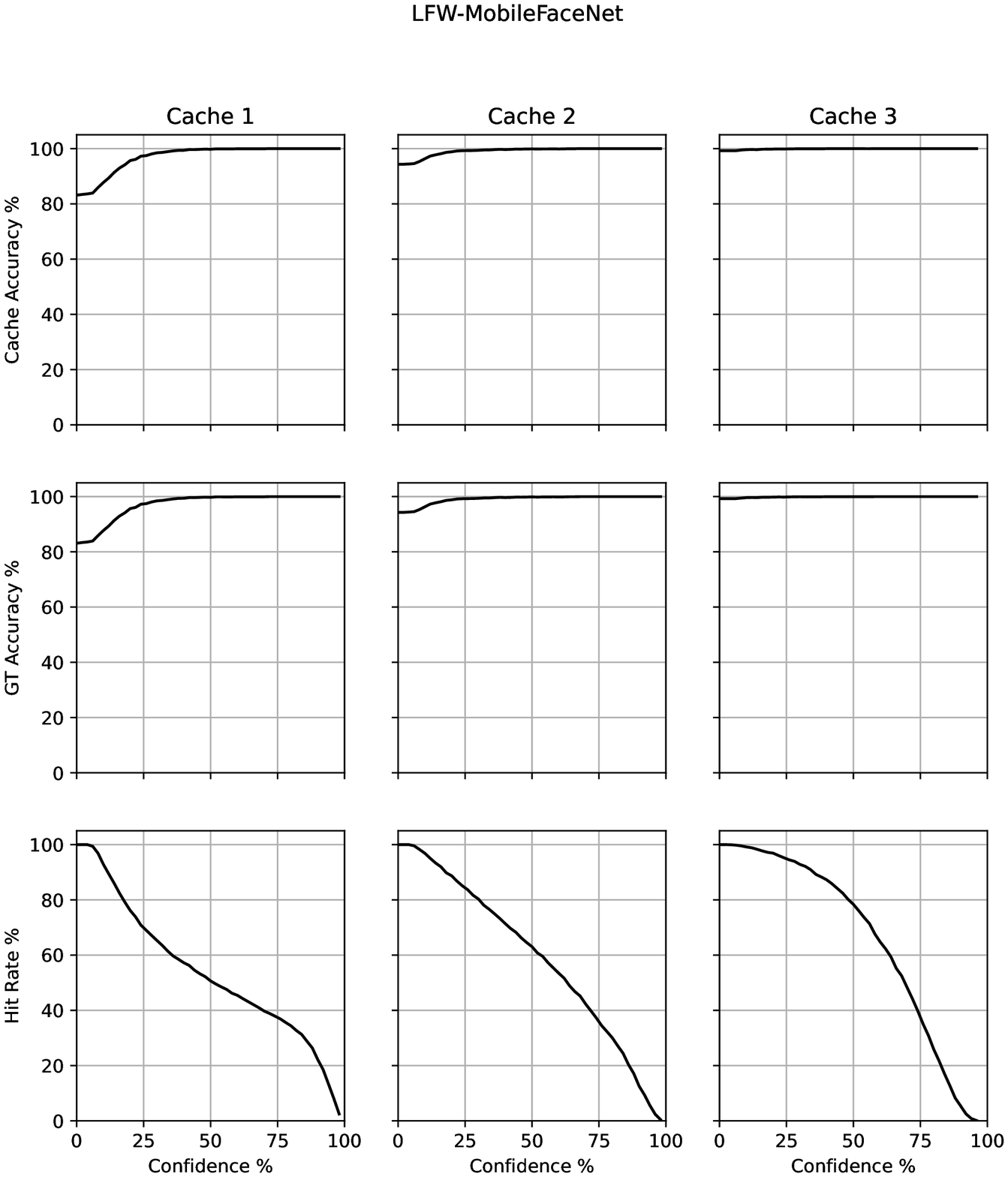}
\caption{Experiment: LFW-MobileFaceNet}
\end{figure}

\begin{figure}[p]
\centering
\includegraphics[width=\columnwidth]{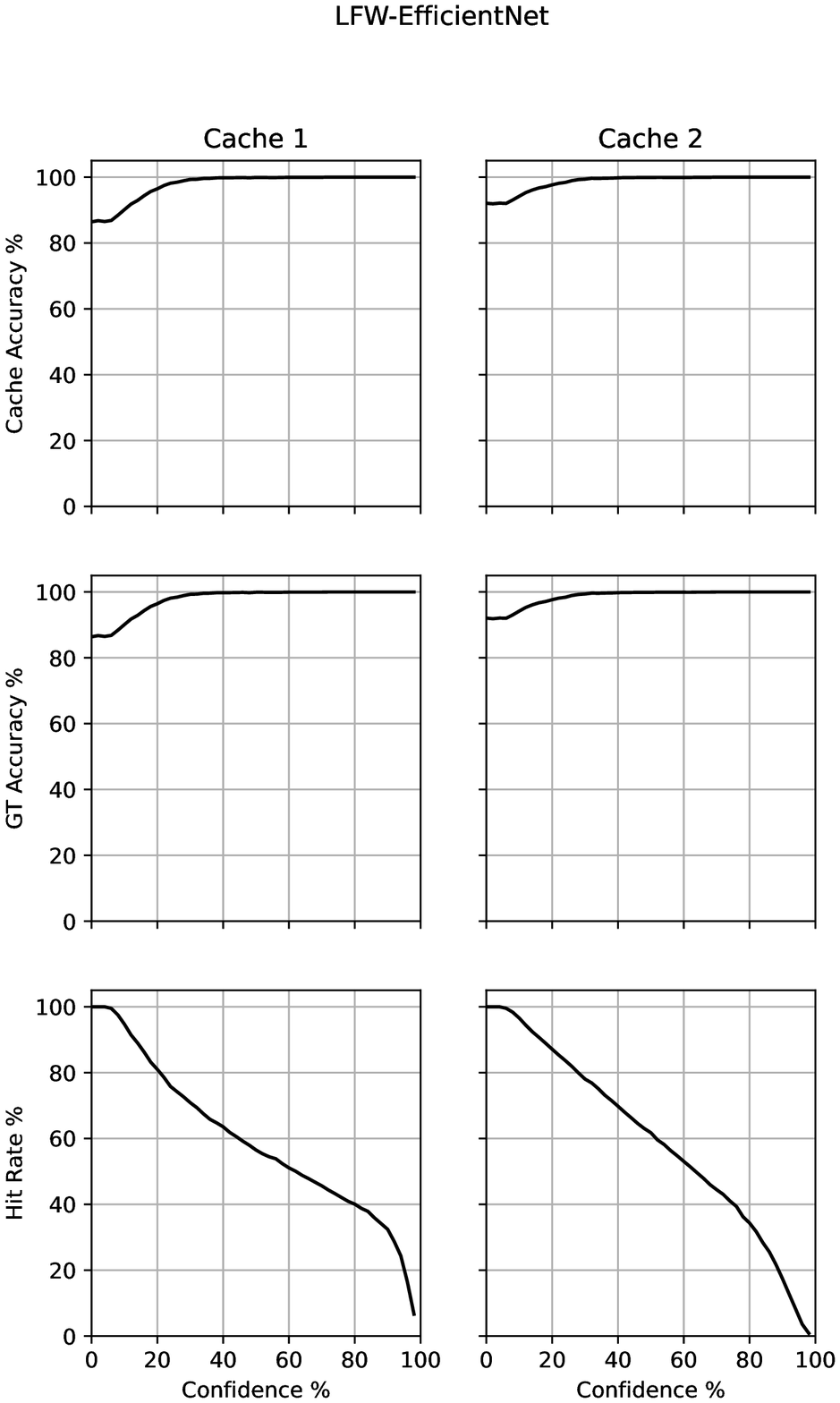}
\caption{Experiment: LFW-EfficientNet}
\end{figure}



\end{appendices}


\bibliography{citations}


\end{document}